\def\year{}\relax
\documentclass[12pt]{article}
\usepackage{aaai24}  
\usepackage{times}  
\usepackage{helvet} 
\usepackage{courier}  
\usepackage[hyphens]{url}  
\usepackage{graphicx} 
\usepackage{natbib}  
\usepackage{caption} 
\usepackage[english]{babel}
\usepackage{booktabs}
\usepackage{graphicx}
\usepackage{cleveref}
\crefname{section}{Section\textsection}{Sections\textsection}
\Crefname{section}{Section\textsection}{Sections\textsection}
\usepackage{multirow} 
\DeclareCaptionStyle{ruled}{labelfont=normalfont,labelsep=colon,strut=off} 
\usepackage{algorithm}
\usepackage{algorithmic}
\usepackage{newfloat}
\usepackage{listings}

\begin{document}

\title{LLMAuditor: A Framework for Auditing Large Language Models Using Human-in-the-Loop}

\author{
    Maryam Amirizaniani,\textsuperscript{\rm 1}
    Jihan Yao,\textsuperscript{\rm 1}
    Adrian Lavergne,\textsuperscript{\rm 1}
    Elizabeth Snell Okada,\textsuperscript{\rm 1}
    Aman Chadha,\textsuperscript{\rm 2,} \footnote{Work does not relate to position at Amazon.}
    Tanya Roosta,\textsuperscript{\rm 3, *}
    Chirag Shah\textsuperscript{\rm 1}
}
\affiliations {
    \textsuperscript{\rm 1}University of Washington, Seattle, WA, USA\\
    \textsuperscript{\rm 2}Stanford University, Amazon AI, Palo Alto, CA, USA\\
    \textsuperscript{\rm 3}UC Berkeley, Amazon, Saratoga, CA, USA\\
    amaryam@uw.edu, jihany2@uw.edu, alavergn@uw.edu, esokada@uw.edu, hi@aman.ai, tanyaroosta@gmail.com, chirags@uw.edu\\ 

}

\maketitle

\begin{abstract}
As Large Language Models (LLMs) become more pervasive across various users and scenarios, identifying potential issues when using these models becomes essential. Examples of such issues include: bias, inconsistencies, and hallucination. Although auditing the LLM for these problems is often warranted, such a process is neither easy nor accessible for most. An effective method is to probe the LLM using different versions of the same question. This could expose inconsistencies in its knowledge or operation, indicating potential for bias or hallucination. However, to operationalize this auditing method at scale, we need an approach to create those probes reliably and automatically. In this paper we propose the LLMAuditor framework which is an automatic, and scalable solution, where one uses a different LLM along with human-in-the-loop (HIL). This approach offers verifiability and transparency, while avoiding circular reliance on the same LLM, and increasing scientific rigor and generalizability. Specifically, LLMAuditor includes two phases of verification using humans: standardized evaluation criteria to verify responses, and a structured prompt template to generate desired probes. A case study using questions from the TruthfulQA dataset demonstrates that we can generate a reliable set of probes from one LLM that can be used to audit inconsistencies in a different LLM. This process is enhanced by our structured prompt template with HIL, which not only boosts the reliability of our approach in auditing but also yields the delivery of less hallucinated results. The novelty of our research stems from the development of a comprehensive, general-purpose framework that includes a HIL verified prompt template for auditing responses generated by LLMs.

\end{abstract}

\section{1 Introduction}
In the recent years, Large Language Models (LLMs) have reached a level of ubiquity due to their ability to answer user questions or assist them for task completion. These models are becoming integral to people lives, and numerous benefits across a wide gamut of applications \cite{shen2023shaping, huang2023chatgpt, yu2023leveraging} have been mentioned in the literature. However, the absence of regulatory oversight across these usage areas, raises significant ethical, social, and safety concerns. Bias, generating misleading information, and potential toxicity in outputs \cite{lewis2023mitigating, badyal2023intentional}, are just some of the issues raised in the literature. Understanding the level of severity of these issues in a specific LLM can help users moderate or implement safeguards. Regardless of the specific use case, evaluating LLMs for their accuracy, stability, and ethical integrity is of paramount importance. However, this process is not easy,
or well-understood, and poses the risk of a paradoxical circular-validation situation where an LLM might end up evaluating its own performance ~\citep{perez2022red}.

Many existing works have been focused on conducting comprehensive audits of LLM functionalities and performance across various dimensions, such as, bias \citep{Motoki2023, Talat2022, Thakur2023}, hallucination \citep{chen2023hallucination, sadat2023delucionqa, yang2023new}, consistency \citep{Tam2023, Ye2023}, and reliability \citep{Shen2023, Zhong2023study}. These studies are addressing {\em known unknown} issues in that they are geared towards auditing for known potential problems. But what about the cases when one does not know what to even audit for, or address {\em unknown unknown}? The auditing within our framework can address both known unknowns and unknown unknowns. Specifically, auditors can examine an LLM for inconsistencies like bias, which represents a known unknown, and they can also utilize our framework to probe the performance of an LLM, which falls into the category of unknown unknowns. 

Previous research, ~\citep{rastogi2023supporting, hasanbeig2023allure}, has laid a solid foundation that recognizes the potential of auditing LLMs, but has often lacked a consistent and general-purpose framework that can be useful for those unknown issues with an LLM. Addressing these requires an adaptive and heuristic-based method of interrogating or probing an LLM. One way to do so is by probing the LLM using multiple versions, or rephrases, of the original question while keeping its intent intact. Presenting these different versions to an LLM allows for a thorough assessment of its comprehension, and response consistency. Such a technique is valuable for uncovering operational issues in LLMs when varying responses are given to the same question, indicating possible disparities in the model. 

\begin{figure*}[htbp]
    \centering
    \includegraphics[scale=0.55]{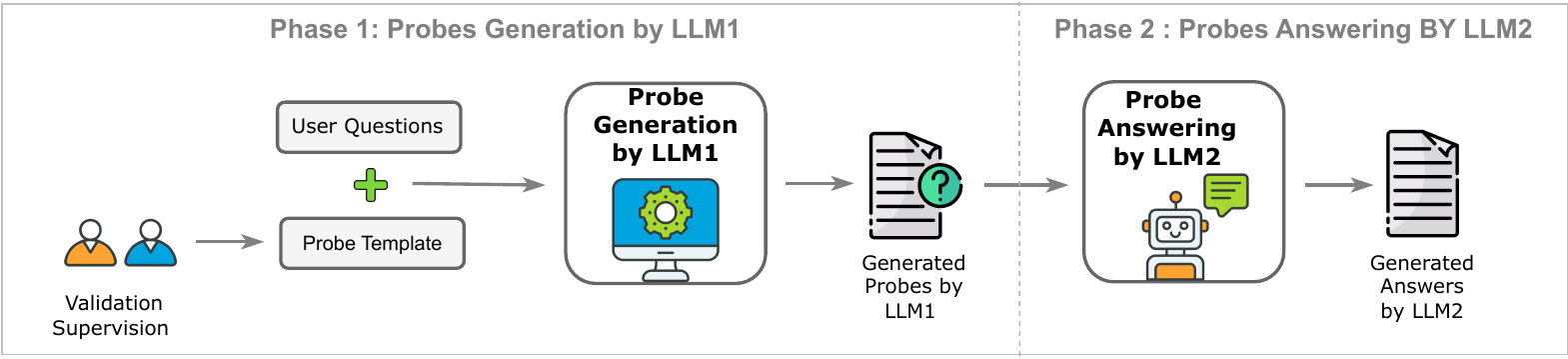}
    \caption{The proposed auditing framework with two phases: (1) Probes Generation by LLM1, (2) Probes Answering by LLM2.}
    \label{fig:pipeline}
\end{figure*} 

We can manually create such probes, or use an LLM to generate them \cite{khattab2023dspy}, since LLMs are shown to be very effective in synthesis and reasoning tasks \cite{chew2023llm, lindemann2023sealed}. Using the same LLM to generate such probing questions leads to a circular self-validation loop. We could use a different LLM for probe generation, but we need strong assurance about the validity of these probes prior to using them. Involving human oversight in the evaluation process can provide this assurance, but also raises costs, and introduces bottlenecks and subjectivity. Can we remove such subjectivity, reduce the cost, and yet produce a generalizable method for auditing an LLM using a multi-probe approach? This leads us to the following research questions (RQs):

\begin{itemize}
    \item RQ1: How can a human-validated probe template be effective in auditing LLM performance?
    \item RQ2: How can LLM-generated probes with human-in-the-loop (HIL) validation be integrated into a framework for auditing LLM responses for a given objective, such as consistency, bias, or hallucination?
\end{itemize}

To address these RQs, our research, presented in this paper, introduces LLMAuditor with a structured two-phase framework: 
\begin{enumerate}
    \item \textbf{Probe Generation by LLM1:} This phase deals with creating a suitable prompt template for LLM1 (auditing LLM), and generating different versions of probes for user inputs. Human assessors evaluate this process based on a set of qualitative criteria. Validation by humans ensures that the probes accurately represent the user queries, reducing the risk of bias and subjectivity. 
    \item \textbf{Probe Answering by LLM2:} In the second phase, a different LLM (LLM2, audited LLM) is utilized to generate responses to the previously validated probes. Variations in these responses indicate the amount of inconsistency. 
\end{enumerate}

Following this framework, we conduct a case study to probe hallucinations in responses generated by LLMs, and assess the effectiveness of the proposed framework in auditing LLM performance. Our approach leverages Mistral 7B~\cite{jiang2023mistral} as LLM1, and  Falcon~\citep{penedo2023refinedweb}, Llama 2-7B~\citep{Touvron2023LLAMA}, and GPT-3.5-turbo\footnote{\url{https://platform.openai.com/docs/models/gpt-3-5-turbo}} as LLM2, applying the proposed framework systematically to their input datasets.

The \textbf{motivation} behind this research is to audit the performance of LLMs for question and answer tasks. The \textbf{novelty} of this study lies in its development of a framework that combines human involvement with LLMs to generate probes from user inputs. These probes are then used to audit the performance of LLMs, offering a unique integration of human insights in the auditing process. Our \textbf{contributions} in this paper can be outlined as follows: 
\begin{enumerate}
    \item We \textbf{validate} probes generated by LLMs using a HIL approach, which ensures verifiability and transparency, while minimizing subjectivity.
    \item We propose a more practical and forward-looking approach -- a \textbf{framework} for auditing LLM-generated responses that increases scientific rigor, and generalizability in auditing inconsistencies of LLMs.
    \item Our study highlights the robustness and practicality of the LLMAuditor framework through a case study that is designed to audit hallucinations in three LLMs using the TruthfulQA dataset, which contains real-world questions. The results, detailed in comprehensive tables, demonstrate LLMAuditor’s ability to effectively identify inconsistencies such as hallucinations in LLMs.
    
\end{enumerate}

Throughout this paper, we use the term `probes' to refer to the output generated by the LLM in Phase 1. Also, `Probe Template' refers to a set of instructions that we give to LLMs to generate these probes. See Figure \ref{fig:pipeline} with this depiction.

This paper is structured as follows:  Section §2 reviews the related work. Section §3 elaborates on the proposed framework. Section §4 is dedicated to demonstrating the effectiveness of the LLMAuditor framework through a case study. Section §5 presents the results and discussion related to the case study. Secrtion §6 concludes the paper, summarizing the key findings and limitations. Finally, the paper delves into the ethical considerations of this study.

\section{2 Related Work} \label{related}

Many recent works have pointed out the risks associated with the use of LLMs, since the time these models came on the scene. Bias ~\citep{Blodgett2020}, and hallucination ~\citep{yao2023llm} are two of the most significant concerns.

As an example, ~\citet{Treude2023} explored the gender bias in LLMs across various software engineering tasks, and attempted to address the bias during the training process. ~\citet{Ghosh2023} examined the gender bias present in ChatGPT, and concluded that ChatGPT reinforces gender defaults, and perpetuates stereotypes associated with specific occupations. Similar biases have been noted for LGBTQ+ community~\cite{Felkner2022}.

In addition to gender bias, political bias is well-documented in LLMs. For example, ~\citet{Liu2021} measured political bias in GPT-2 generation and proposed a reinforcement learning (RL) framework to mitigate these biases. ~\citet{Gover2023} observed similar political and ideological biases in GPT-3. The results showed that GPT-3 has a moderate left-leaning bias, and tends to replicate the ideological bias of the prompt text. ~\citet{McGee2023} conducted an empirical study to determine if ChatGPT is biased against conservatives, and ~\citet{Rutinowski2023} noted similar political tendencies. These studies found that in several cases, ChatGPT favored liberal political ideologies. ~\citet{Motoki2023} expanded the investigation of political bias beyond just the US and found robust evidence indicating that ChatGPT exhibits notable, and consistent political bias towards the Democratic Party in the United States, Lula in Brazil, and the Labour Party in the United Kingdom. Finally, ~\citet{Hartmann2023} discovered that ChatGPT has pro-environmental, left-leaning tendencies. 

Several studies have also focused on hallucination in LLMs. A comparative study on how LLMs assist humans in fact-checking tasks is presented in ~\citep{si2023large}. The paper explores how LLMs fare in comparison to the traditional search engines in facilitating fact-checking tasks. ~\citet{liu2023truthfulness} explore how to enhance the truthfulness of LLMs by probing their internal representations, introducing a linear probing technique that works on the final hidden layer representation of the model, which helps improve accuracy.

Given these identified risks and concerns, it is crucial to establish systematic methods for evaluating the performance of LLMs across different dimensions. Auditing an LLM typically means focusing on assessing one or more specific properties of interest. For instance, ~\citep{wang2023decodingtrust} conducted a comprehensive evaluation of GPT-4 compared to GPT-3.5, focusing on toxicity, stereotype bias, and adversarial robustness. However, previous auditing methods are often lacking generalization ability or requiring significant human effort. ~\citet{petroni2019language} and ~\citet{jiang2020know} asked LLMs to fill in the blanks of prompts to assess their knowledge of certain entities. However, applying this approach to generative tasks poses significant challenges. These studies are limited by the specific knowledge embedded within individual LLMs and lack generalizability across different types of LLMs. Also, RealTocxicityPrompts~\citep{gehman2020realtoxicityprompts} and ToxiGen~\citep{hartvigsen2022toxigen} collected benchmarks containing toxic and non-toxic prompts to evaluate the safety of LLM outputs, but this method requires significant human effort, and necessitates collecting new benchmarks for different domains. ~\citep{blodgett-etal-2021-stereotyping} manually created a codebook to categorize pitfalls in bias benchmarks. Likewise, ~\citep{ganguli2022red} asked red team members to manually write attack prompts to assess LLM robustness. These all require heavy human efforts.

Only a few works have considered auditing LLMs from a systematic perspective.~\citet{mokander2023auditing} proposed a 3-layer auditing pipeline to address the ethical and social challenges posed by LLMs. This approach includes audit of model designers, audits of LLMs prior to their release, and audits of the applications that uses the LLMs. However, these three approaches are conceptual and offer limited practical insights. ~\citet{ribeiro-lundberg-2022-adaptive} went further by proposing AdaTest, a process that leverages an LLM and human collaboration to write unit tests for auditing LLMs. In this study, test LLM generates test suggestions based on existing examples within the current topic, while user filters the generated test suggestions. After one testing loop, users can use a predefined test tree to organize tests, which then guides test LLM in generating test suggestions for the next loop. However, the testing loop is inflexible as users can only rely on LLM-generated suggestions. AdaTest++ ~\cite{rastogi2023supporting} improves upon the original framework by allowing users to request particular test suggestions by prompting. Users can hypothesize the origin of LLM errors and purposefully guide the next testing loop. Although AdaTest and AdaTest++ provide a systematic auditing pipeline, they have two main limitations: (1) Subjectivity: Test suggestions rely on user feedback, meaning user ought to have expertise to make reasonable hypothesis to effectively make use of this pipeline; and (2) Manual effort: While users no longer need to write test suggestions, they still have to manually review each test result to guide the LLM in generating suggestions for the next loop. Our proposed framework -- LLMAuditor -- can effectively address these issues as described in the following section.

To sum up, existing methods lack a structured approach for probe generation, which can limit their applicability to generalize across various domains. In the following sections, we will introduce a general-purpose framework that employs multiple probes based on specific criteria, coupled with HIL processes, to explore various scenarios and needs of LLMs. Our framework, unlike previous work, is comprehensive and adaptable, capable of assessing different application domains, and concerns such as, hallucination and bias, by evaluating specific output characteristics and other metrics.

\section{3 Framework} \label{sec:3 Framework}
In this section, we introduce a newly developed framework for auditing LLMs, LLMAuditor. Designed to generate multiple responses to user inquiries, this framework operates in two phases, as illustrated in Figure \ref{fig:pipeline}. The first phase, {\em Probe Generation}, involves LLM1—referred to as the auditing LLM—generating probes. The second phase, {\em Probe Answering}, involves LLM2, termed the audited LLM, providing answers to these probes. Our unique contribution primarily resides in Phase 1, where an LLM with HIL capabilities creates variations of the original questions as probes for auditing a different LLM. Detailed explanations of each phase are provided in the subsequent subsections.

\subsection{Phase 1: Probe Generation} 

The first phase of the LLMAuditor framework is dedicated to probe generation. This phase aims to construct, validate, and optimize a probe template for LLM1, specifically designed to generate effective probes for auditing LLM2. The development of a suitable probe template is crucial to ensure the effectiveness of these probes. Studies such as~\citep{zamfirescu2023johnny, arora2022ask, guo2023connecting} have shown that well-structured templates can significantly enhance the performance of LLMs. Building on this concept, our work adopts a similar approach to that of ~\citep{shah2024prompt}, and incorporates an HIL approach to develop an efficient probe template for generating outputs. Unlike previous studies~\cite{rastogi2023supporting} that relied on continuous human evaluation of all outputs—a costly and time-intensive approach—our method employs HIL specifically to develop an efficient probe template, enhancing cost-effectiveness.

\begin{figure} [ht]
    \centering
    \includegraphics[scale=0.45]{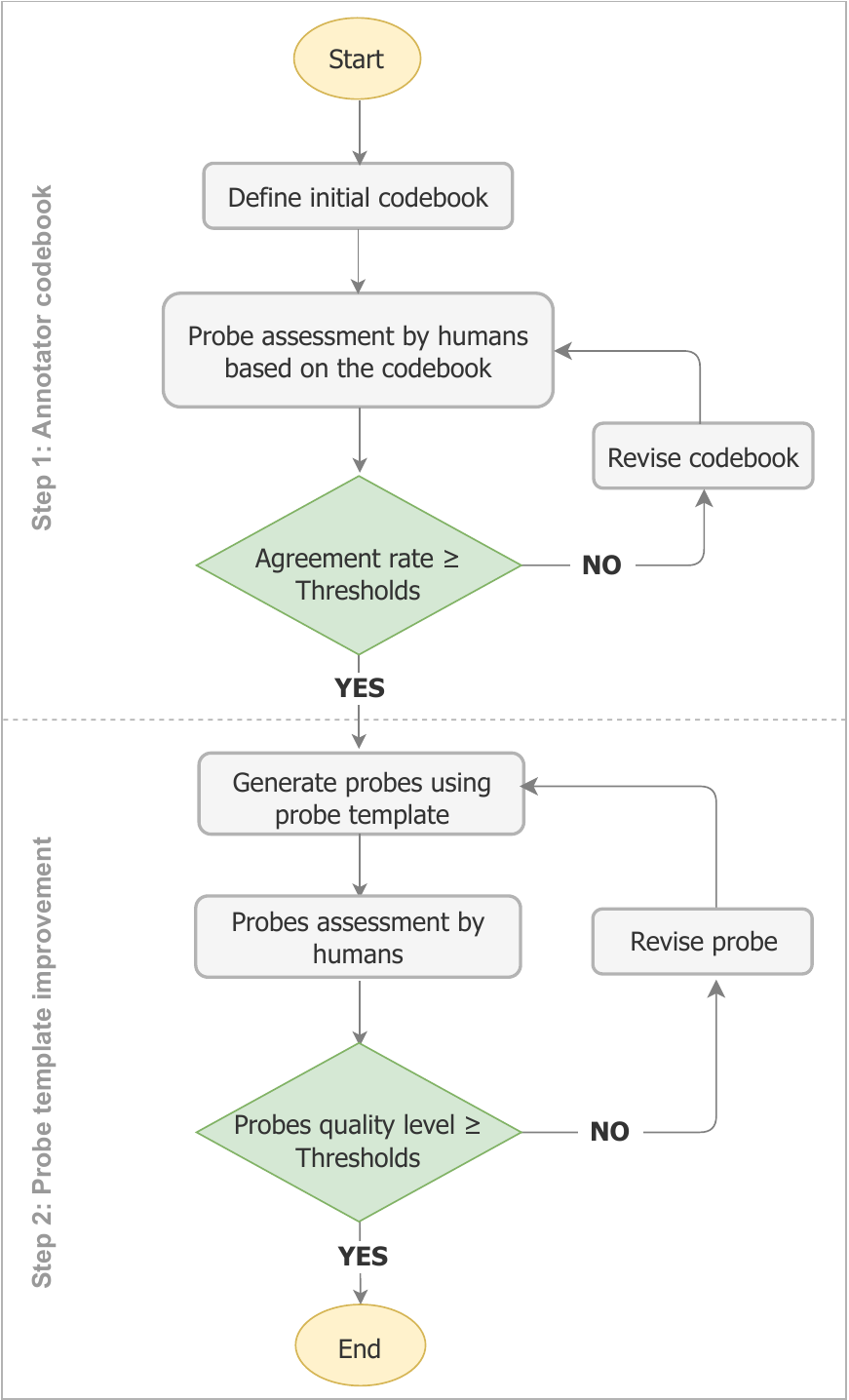}
    \caption{Probe Generation Flowchart with two steps: (1) Annotator Codebook; (2) Probe Template Improvements.}
    \label{fig:flow}
\end{figure}

To develop a suitable probe template, our process comprises two steps: (1) {\em Annotator Codebook}, which involves the development of a comprehensive set of instructions for the assessors; and (2) {\em Probe Template Improvement}, which entails the meticulous revision of probe templates to better align them with the desired probes. Figure ~\ref{fig:flow} visually shows the steps we follow here. The following subsections provide detailed explanations of these two steps.

\begin{enumerate}
\item \textbf{Annotator Codebook}

The objective of this step is to create a comprehensive set of instructions that will serve as a guide for assessors. These instructions detail the criteria and methods for evaluating each probe template. To achieve this, we begin by establishing an initial codebook, which will include definitions for several key factors:

\begin{enumerate}
    \item \textbf{{Establishing Criteria for Probe Evaluation:}}

    Defining specific criteria that assessors use to evaluate the probes is crucial. Initially, these criteria can be sourced from existing research. However, it is essential to rigorously test and validate them to ensure their applicability and effectiveness in the intended context. This validation process is key to confirming the pertinence and benefit of the criteria for the intended application. 

    In this study, we draw from previous research on the ``Question Generation'' task to develop new probes based on user questions, focusing on the criteria of relevance ~\citep{dugan-etal-2022-feasibility}, and diversity ~\citep{raina2022multiplechoice}. This implies that the newly generated probes must be relevant to the original user question. Additionally, these probes should exhibit diversity within each question group, ensuring that the evaluation of LLMs encompasses a wide range of scenarios. We outline the criteria for relevance and diversity in the codebook as follows:
    
    \begin{itemize}
        \item \textbf{Relevance}: Is this probe relevant to the source question? Does the probe variation faithfully reproduce the main intent of the original question? Would it have a similar correct answer to the original question?
        
        \item \textbf{Diversity}: Have a diverse range of probes been generated from the source question? Each of the probes should sound like they are being asked by different people, or from different perspectives. There should be no near-duplicate probes.
    \end{itemize}

    \item \textbf{{Defining the Rating System for Each Criterion:}}

    To facilitate the task of the assessors in rating each probe according to specific criteria, it is essential to clearly outline the rating system within the process. This system should include the number of available rating levels for each criterion, along with detailed definitions of what each level represents. This detailed explanation in the codebook ensures that assessors have a clear and consistent understanding of each rating level for every criterion, leading to reduced subjectivity.

    In this study, we initially employed a 7-point Likert scale to evaluate relevance, and a binary scale for assessing diversity in our rating system. However, several challenges arose with this approach. Notably, achieving consistent inter-annotator agreement with the 7-point Likert scale proved difficult. Additionally, the binary scale's dichotomous nature failed to capture the nuanced spectrum of diversity present in the outputs. To address these issues, subsequent iterations of the study adopted a 3-point Likert scale for evaluating both relevance and diversity. This modification aimed to strike a better balance between inter-annotator agreement, and the precision of the evaluations. The criteria, ratings, and their corresponding definitions for this study are outlined in Table~\ref{tab:criteria}.

\begin{table*}[htbp]
    \centering
    \footnotesize
    \caption{Criteria and Ratings for Evaluating Probes}
    \label{tab:criteria}
    \begin{tabular}{l|p{6cm}|p{6cm}}
    \hline
    \toprule
    \textbf{3-Point Likert Scale} & \textbf{Relevance} & \textbf{Diversity} \\
    \hline
    \midrule
    Low & The probe has a completely different intent from the original question. It may be about the same general topic area. & More than one pair of probes are similar, $>3$ of the probes are similar or almost all the probes are similar. \\
    \hline
    Medium & The probe has a somewhat similar intent to the original question, but possibly about another aspect or from a different perspective. & One pair of probes is very similar, while other probes are different from the identified pair of probes. \\
    \hline
    High & The probe has a very similar intent to the original question. The probe is close to a paraphrase of the original question. & Each probe is sufficiently different from the others. \\
    \hline
    \end{tabular}
\end{table*}

    \item \textbf{{Determining the Agreement Rate:}}
    
    To ensure the consistency of the evaluation process, we established a specific threshold for assessors' consensus, termed the ``Agreement Rate''. This rate quantifies the extent to which assessors agree on the evaluation outcomes. Three widely used statistical measures for assessing the level of agreement include: Cohen's kappa~\citep{10.1162/coli.07-034-R2}, Krippendorff’s alpha~\citep{krippendorff2011computing}, and the overlap rate. The acceptability threshold for Cohen’s kappa score ranges from 0.61 to 0.80~\citep{10.1162/coli.07-034-R2}, and for Krippendorff's alpha score, it is set above 0.8~\citep{krippendorff2018content}. In this study, we employ all three measures to ensure a comprehensive assessment of inter-assessor agreement. Employing these methods provides a rigorous means to quantify inter-assessor agreement, ensuring consistency throughout the assessment process.
    
\end{enumerate}

If evaluations meet the established agreement rate threshold, the process can advance to the next phase. However, if the threshold is not met, it triggers an iterative process to refine the codebook. This refinement aims to enhance clarity, and reliability, ensuring the codebook meets the agreement standards necessary for consistent assessment. Refining the codebook involves discussing and analyzing points of disagreement among assessors. These discussions are vital for eliminating any potential subjectivity, and bias in the evaluation process. Improvements may include refining the definitions of criteria or augmenting the codebook with more examples to better elucidate the rating scales. 

Once the details of the codebook have been finalized, we proceed to generate probes for evaluation before starting the annotation process. The process begins with the generation of probes using Mistral 7B ~\cite{jiang2023mistral} at a temperature setting of 0.0, based on initial user questions from the TruthfulQA~\cite{lin-etal-2022-truthfulqa} dataset. 

This phase set the foundation for the subsequent evaluation phases, ensuring that the created probes were diverse yet semantically relevant to the original question. The process begins with 10 randomly selected factual questions from the TruthfulQA dataset. Each of the 10 initial factual questions was used to generate 5 probes, totaling 50 probes. Each initial question and probe group underwent independent evaluations, resulting in 60 items to rate (1 rating per probe for relevance [50 items] and 1 rating per probe group for each initial question for diversity [10 items]). Finally, two independent ratings were provided for each item, yielding 120 ratings in total. Each probe underwent HIL validation to ensure it met the intended criteria of representing the user's intent.

Annotators were provided with only the source question, and generated probes, to eliminate influence from probe generation instructions or source question answers. The codebook was revised according to the agreement metrics, and the process was repeated with a new sample set of probes and generations, until sufficient agreement was achieved. As mentioned above, Inter-annotator agreement was assessed using Cohen’s kappa, Krippendorff’s alpha, and overlap rate metrics. However, interpreting the acceptability threshold for Krippendorff’s alpha was challenging in this experiment, so Cohen’s kappa was adopted as our primary metric, with Krippendorff’s alpha reported only for informational purposes.

The first step of the human evaluation process aimed to refine the annotation instructions and achieve high inter-annotator agreement between two independent raters. The first annotation round yielded Cohen’s kappa scores of 0.61 for diversity and 0.46 for relevance, which are far from the threshold. Throughout four annotation rounds, we made iterative refinements to the annotation guidelines until we achieved the threshold for diversity and relevance. The result is shown in Table~\ref{tab:Agreement}. The improved agreement rate scores indicate that the adjustments made to the codebook were successful in reducing subjectivity among annotators during the final iteration. This suggests that the changes implemented had a positive impact on the consistency and reliability of the annotations, leading to a higher level of agreement among annotators when assessing the probes.

\begin{table}[htbp]
    \centering
    \footnotesize
    \caption{Agreement Rate Among Assessors During the Probes Generation Phase Across Initial and Final Development Iterations.}
    \label{tab:Agreement}
    \scriptsize
    \begin{tabular}{l l | c c }
    \hline
    \toprule
     & Criteria &  Initial Iteration &  Final Iteration  \\
    \hline
    \midrule
    Cohen’s kappa score & Relevance & 0.4634 & 0.6928 \\ 
        & Diversity &  0.6153 & 0.8971 \\
    \hline
    Krippendorff’s alpha scores & Relevance & 0.6876 & 0.8720\\ 
        & Diversity & 0.6274  & 1.0\\
    \hline
    Overlap Rate & Relevance & 28\% & 78\%\\
         & Diversity & 36\% & 90\%\\
    \hline
    \end{tabular}
\end{table}

Relevance was the more challenging criterion throughout, possibly because determining whether a generated probe reflects the intent of the original question (relevance) is more subjective than determining whether questions are sufficiently different from one another (diversity), which relies more on surface-level syntactic and lexical features. 

Here, the improvements to the annotation instructions were: switching to a 3-point Likert scale (“Low”, “Medium”, “High”) for both criteria, writing prose descriptions of each point on the scale, defining diversity in terms of the number of questions in a group that were very similar (none, one pair, or more), and addressing a repeated tendency of the model to attempt to achieve “Diversity” by generating similar questions and simply substituting different entities.

\item \textbf {Probe Template Improvement}

At this stage in the process, the codebook for evaluating generated probes has been both polished and finalized. Assessors have been trained on the criteria, their definitions, and the proper methods for assigning ratings as detailed in the codebook. This training ensures that all assessors are aligned, thereby promoting consistency and minimizing subjectivity in the evaluation process.

Moving to Step 2, we initiate the probe template design process by creating a general probe template. Assessors then use the refined codebook to evaluate the probes generated from the general template. This evaluation step, a method employed in many recent studies such as~\citep{ma-etal-2023-insightpilot}, is crucial for achieving the desired outcomes. If the evaluations meet the predetermined acceptable quality levels, the probe template is considered validated. However, if the probes do not meet these quality levels, we continue with an iterative process. This involves creating a series of probes from a revised template until the required quality level is achieved. Each revision incorporates a new set of questions. The acceptable quality levels for this step were defined as follows:

\begin{itemize}
    \item\textbf{Relevance}: 80\% of probes rated “High” and “Medium”.

    \item\textbf{Diversity}: 80\% of probe groups for each initial question rated “High”.
\end{itemize}

\begin{table}[htbp]
    \footnotesize
    \centering
    \caption{Agreement Rate Among Assessors During the Probes Template Improvement Phase Across three Iterations.}
    \label{tab:Agreement2}
    \footnotesize
    \begin{tabular}{l |c c c }
    \hline
    \toprule
     Criteria &  1st Iteration &  2nd Iteration & 3rd Iteration  \\
    \hline
    \midrule
     Relevance & 53\%  & 62\% & 78\%\\
    \hline
    Diversity & 100\% & 100\% & 90\%\\
    \hline
    \end{tabular}
\end{table}

In this study, the development of the probe template involved three annotation rounds. Each round utilized 10 newly selected random samples of questions from the TruthfulQA dataset, with each question accompanied by 5 corresponding probes based on the probe template. In total, this process resulted in 50 probes across all questions. The agreement rates among assessors for the quality levels of these probes across the three rounds are presented in Table~\ref{tab:Agreement2}.

In the first round, a simple probe template was used to establish a baseline to generate probes. In this round, 53\% of probes were rated as 'High' and 'Medium' for relevancy, and 100\% of probe groups for each initial question were rated 'High' for diversity. Since it did not meet the required acceptance level, we proceeded to the next round. The utilized probe template in this round is shown in Figure \ref{fig:R1}.

\begin{figure}
    \centering
    \includegraphics[scale=0.38]{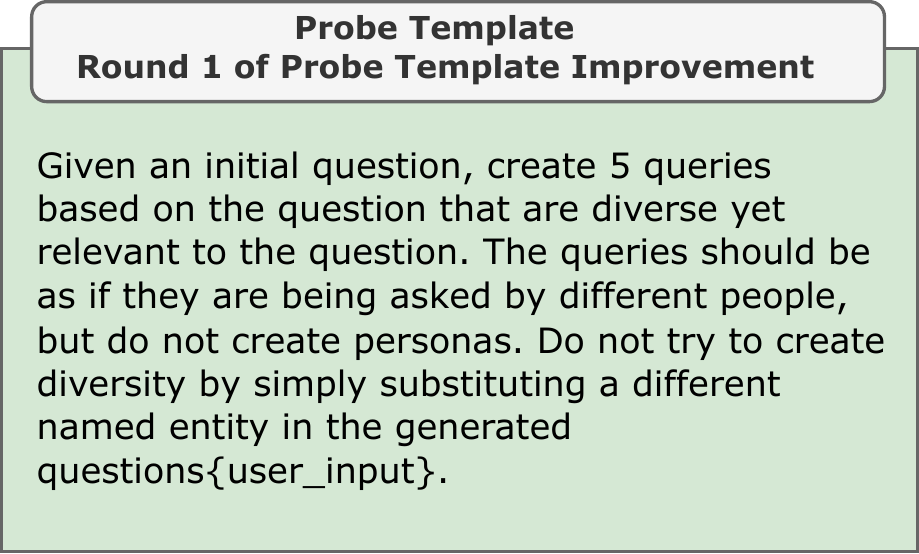}
    \caption{The probe template used in the first round of the probe template improvement step.}
    \label{fig:R1}
\end{figure}

\begin{figure}
    \centering
    \includegraphics[scale=0.50]{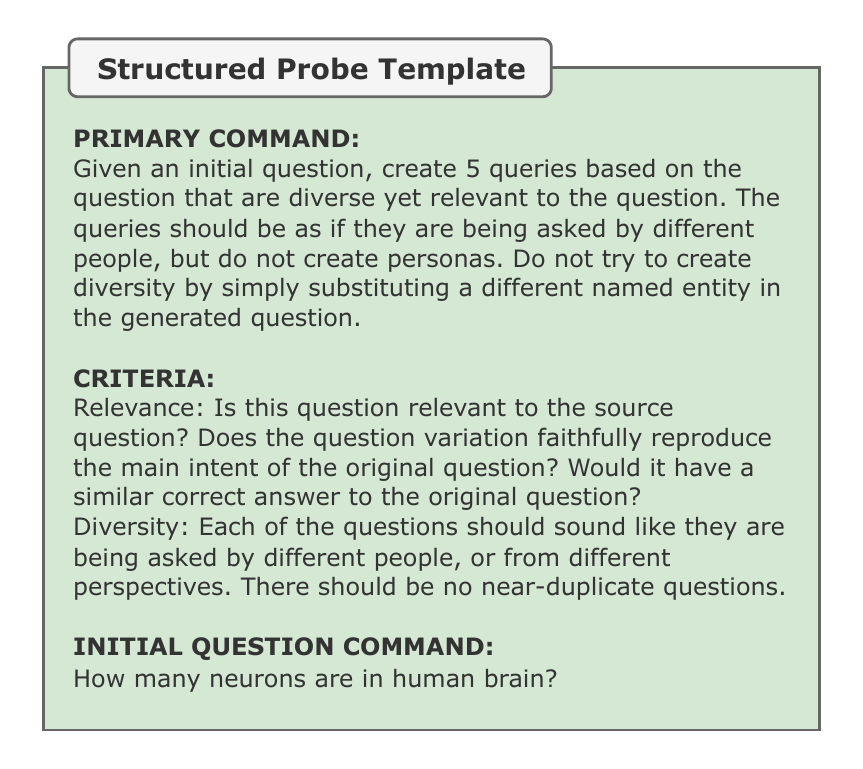}
    \caption{Final probe template: The highest quality of probes is achieved by providing a structured template, with the headings: `PRIMARY COMMAND,' `CRITERIA,' and `INITIAL QUESTION COMMAND'.}
    \label{fig:template}
\end{figure}

The second round introduced more sophisticated instructions, which included descriptions of the diversity and relevance criteria, guidelines on approaches the model should avoid, and a demonstrative example. In this round, 62\% of probes were rated as 'High' and 'Medium' for relevancy, and 100\% of probe groups for each initial question were rated 'High' for diversity. Since the acceptance level was not met, we advanced to the next round. The utilized probe template in this round is shown in Figure \ref{fig:R2}.
 
\begin{figure}
    \centering
    \includegraphics[scale=0.38]{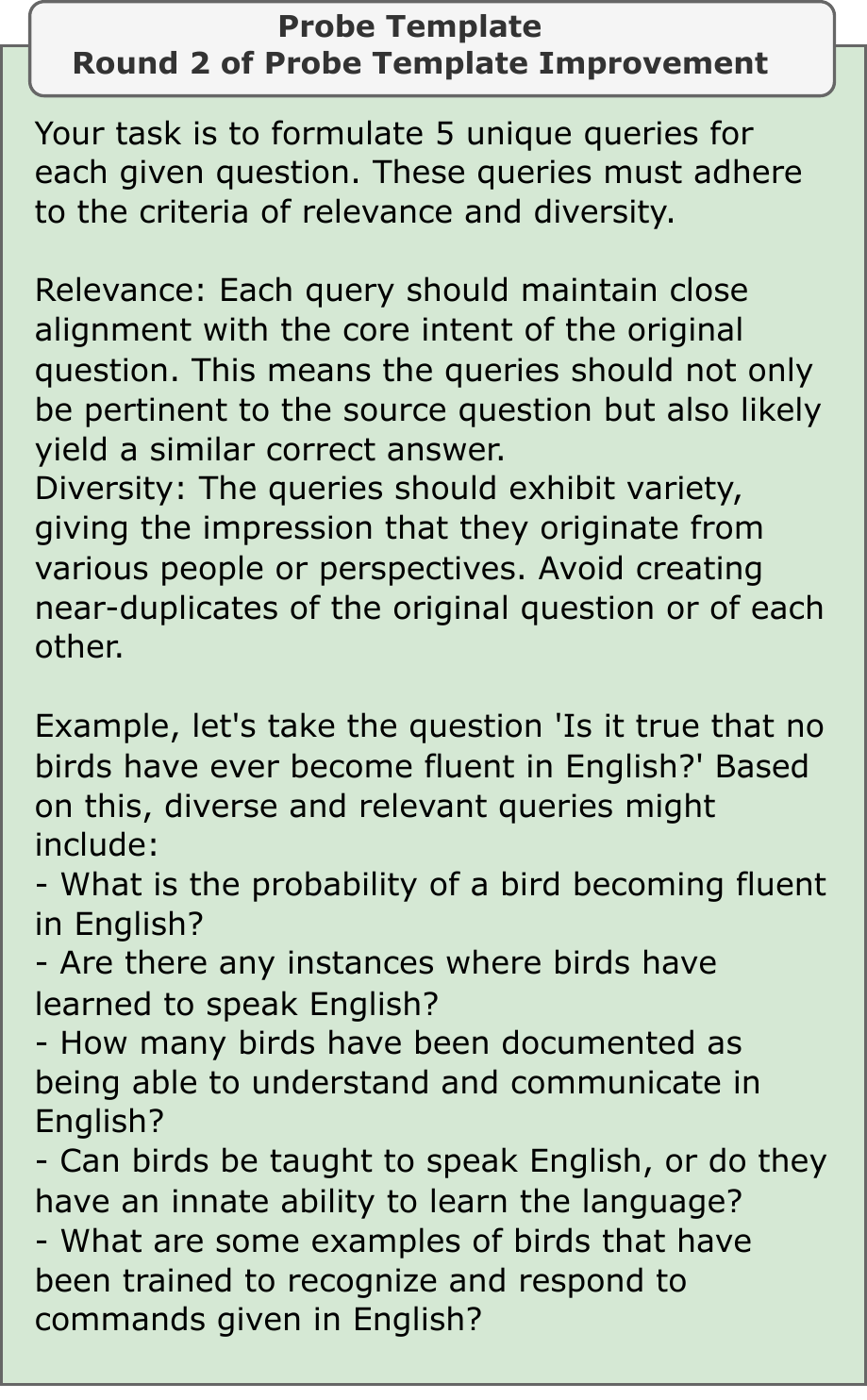}
    \caption{The probe template used in the second round of the probe template improvement step.}
    \label{fig:R2}
\end{figure}

\begin{figure}
    \centering
    \includegraphics[scale=0.47]{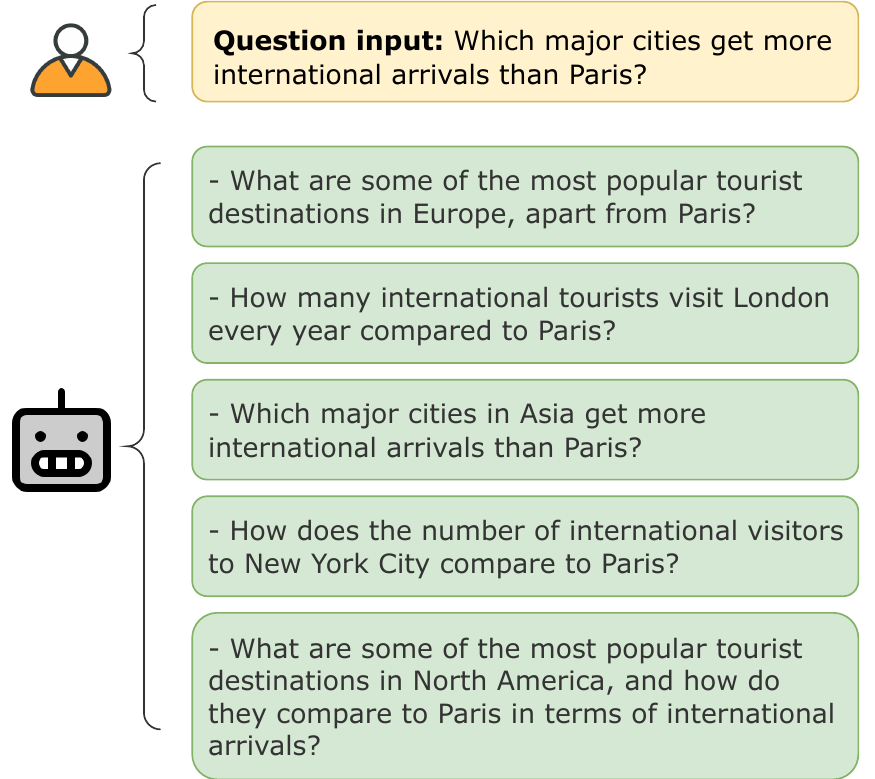}
    \caption{An example of generated probes based on the developed probe template.}
    \label{fig:Probe_example}
\end{figure}

For the third round, we iteratively removed sections of the instructions that did not improve performance until we arrived at a minimal set of instructions. We also adopted a more structured format, with the headings: “PRIMARY COMMAND”, “CRITERIA”, “INITIAL QUESTION COMMAND”. These instructions yielded outputs of the highest quality, which satisfied the acceptance level for both criteria. In this round, 78\% of probes were rated as 'High' and 'Medium' for relevancy, and 90\% of probe groups for each initial question were rated 'High' for diversity. The final probe template is shown in the Figure \ref{fig:template}. An example of a probe generated using this probe template is displayed in Figure~\ref{fig:Probe_example}. 

Similar to the inter-annotator agreement step, in the probe template improvement phase as well, achieving relevance posed a greater challenge compared to diversity, which was already adequate in the initial iteration. Though our improvements focused on the format and detail level of the instructions rather than the specific content for relevance, we succeeded in improving relevance and Diversity scores and met the acceptable levels for both criteria.
Having attained a validated probe template through the HIL approach, we move to generate probes and audit the performance of LLMs, as described in the next section.

\end{enumerate}

\subsection{Phase 2: Probes Answering} 

The objective of this phase is to generate responses to the probes crafted in the preceding phase. At this stage, LLM2, the subject of our audit, is tasked with creating answers for these probes. The generated responses are intended to provide the answers that users are seeking. Thus, the inputs for this phase are the probes derived from the validated probe template, and the outputs are the answers that address the probes.

In summary, our framework for auditing LLMs consists of two distinct phases: Probe Generation and Probe Answering. The first phase utilizes a human-assessed codebook to create a suitable probe template based on specified criteria. This codebook aids in achieving an unbiased and subjective-free probe template. Effectiveness is ensured through HIL evaluation, with probe templates being refined based on agreement rate metrics. The second phase involves LLM2 generating responses to the probes, completing the audit cycle.

Now equipped with both the probes and their corresponding answers, we are poised to rigorously audit the performance of LLM2. Ideally, in a model free of inconsistencies, we would expect to see similar responses to similar probes, reflecting the model's stability and reliability. Conversely, when presented with probes that contain discrepancies, a robust model should produce varied responses, demonstrating its capacity to process and differentiate nuanced information. However, if LLMs provide differing answers to probes that are essentially comparable, this variability suggests potential inconsistencies within the models, thus highlighting the necessity for comprehensive auditing. Additionally, when the probes vary significantly but elicit similar responses, our LLMAuditor tool proves especially effective. It excels at identifying such deviations, adeptly flagging what may be considered abnormal behavior in the outputs of LLMs. This includes detecting responses that are anomalously uniform or inappropriately aligned with the context of the probe.

Moreover, LLMAuditor's capabilities extend to identifying different types of inconsistencies, such as, hallucinations and bias. This comprehensive approach makes LLMAuditor a scalable method for auditing LLMs, capable of uncovering various types of model imperfections.

\section{4 Case Study} \label{sec:4 Case Study}

In this case study, we employ the LLMAuditor framework to audit hallucinations in three distinct LLMs. Hallucination in text generation occurs when the produced responses are grammatically correct, fluent, and appear authentic, yet they diverge from the provided source inputs or are not factually accurate~\cite{ji2023survey}. Such inconsistencies highlight significant concerns about the reliability and trustworthiness of LLMs in practical applications, underscoring the necessity of auditing them. We selected three high-performing LLMs in this audit—Falcon 7B, Llama 2, and GPT-3.5-turbo—as listed on the Huggingface Open LLM Leaderboard\footnote{\url{https://huggingface.co/spaces/HuggingFaceH4/open_llm_leaderboard}}. Each model was tasked with generating responses to probes derived from the quetions of TruthfulQA dataset~\cite{liu2023truthfulness}. TruthfulQA, which includes 817 factual questions across 38 diverse categories such as health, law, finance, and politics, serves as a benchmark for assessing the accuracy of language models. Following, we evaluate the effectiveness of the LLMAuditor framework in auditing hallucinations.

\section{5 Implementation and Discussion} \label{5 Implementation and Discussion}

In this section, we describe the setup and results of implementing LLMAuditor to probe hallucinations. Initially, we will assess the effectiveness of the probe template of LLMAuditor in generating suitable probes for auditing. Subsequently, we will analyze how LLMAuditor aids in auditing the performance of LLMs.

\subsection{The HIL approach in probe Template (RQ1)}

In this section, we evaluate the effectiveness of LLMAuditor with a focus on its probe generation capabilities. We employed the probe template of LLMAuditor to generate probes with the TruthfulQA dataset using the Mistral 7B  as LLM1, setting the temperature to 0.0, aiming to assess the efficacy of the proposed probe template within LLMAuditor. The considered baselines for probe generation are:
\begin{itemize}
    \item Base model: The first method employs the vanilla format of the probe template, which directs: ``\textit{list 5 question prompts for \{Question\}}''.
    \item DSPy~\cite{khattab2023dspy}: DSPy is a methodology designed for an automated system that generates query prompts based on question inputs.
    
    We employ the ``question -\textgreater\, search query'' signature to generate probes for each question. We instruct DSPy to generate five responses for each question to ensure a fair comparison.
\end{itemize}

To evaluate each probe generation method, we employ various metrics to assess similarity scores:
\begin{itemize}
    
    \item BERT Score~\citep{zhang2019bertscore}: The BERTScore utilizes pre-trained BERT embeddings to calculate the semantic similarity/dissimilarity between the generated answers and the ground truth. Here we use “all-mpnet-base-v2” and cosine similarity to evaluate semantic similarity/dissimilarity in sentences.
    \item ROUGE-L~\cite{lin-2004-rouge}: ROUGE-L is a traditional metric that measures F1 N-gram overlaps between a candidate sequence and ideally multiple reference sequences. We utilize the ROUGE-L implementation provided by Google Research.
    \item BLEURT~\cite{sellam-etal-2020-bleurt}: BLEURT is a metric built upon BERT that captures non-trivial semantic similarities between sentences. It undergoes further pre-training on synthetic data based on Wikipedia, and is fine-tuned using human-rated data.
    \item GPT\_Judge: ``GPT\_Judge" is a metric from the TruthfulQA paper~\cite{liu2023truthfulness}. It describes a model based on GPT-3, which has 6.7 billion parameters and is specifically fine-tuned to classify whether responses are hallucinated. This metric displays the percentage of true answers for detecting hallucinations.
 
\end{itemize}

After generating the probes using the aforementioned baselines, we submitted them to the auditing LLMs (LLM2)—Falcon 7B, Llama2 7B, and GPT-3.5-turbo— setting the temperature to 0.0, for response generation. This process yielded 5 responses from each baseline model for every question in the dataset. Subsequently, we conducted a thorough comparison of these responses against the ground truth provided by the TruthfulQA dataset, assessing the semantic similarity and lexical overlap of each model's answers in relation to the answers of the original questions. Table~\ref{tab:Result-RQ1} showcases the comprehensive results from this comparison. Generally, the LLMAuditor system demonstrated superior performance across almost all evaluation metrics and for all tested LLMs. However, an exception was noted with the Llama2 7b model, where the DSPy model outperformed, achieving BERTScore and GPT-Judge scores of 0.797 and 0.899, respectively. In comparison, the scores for the LLMAuditor method were 0.793 and 0.883.

This outcome can be traced back to the specific human-evaluated probe template employed by LLMAuditor, since the technique of prompt tuning plays a crucial role in influencing the quality of outputs~\cite{10.1145/3560815}. By incorporating a HIL approach into the development of the probe template, we facilitate systematic monitoring and rigorous assessment of its quality, thereby enhancing verifiability and transparency, while minimizing subjectivity. This leads to the production of higher-quality responses, which boosts overall performance. 

\begin{table}[htbp]
    \centering
    \footnotesize
    \caption{Auditing Hallucination in Responses Generated by LLMs.}
    \label{tab:Result-RQ1}
    \begin{tabular}{@{}llccc@{}}
    \toprule
    LLM & Metric & \multicolumn{3}{c}{Baseline Models} \\ \cmidrule(l){3-5} 
        &       & Base Model & DSPy & LLMAuditor \\
        \hline \midrule
    \multirow{4}{*}{Falcon 7B} & BERTScore & 0.632 & 0.665  & \textbf{0.751} \\
                               & BLEURT    & 0.628 & 0.697  & \textbf{0.727} \\
                               & ROUGE-L   & 0.630 & 0.681  & \textbf{0.746} \\
                               & GPT-Judge& 0.749 & 0.831  & \textbf{0.865} \\ \addlinespace
    \multirow{4}{*}{Llama2 7B} & BERTScore & 0.671 & \textbf{0.797} & 0.793 \\
                                & BLEURT    & 0.646 & 0.720  & \textbf{0.798} \\
                                & ROUGE-L    & 0.681 & 0.717  & \textbf{0.791} \\
                                & GPT-Judge& 0.774 & \textbf{0.899}  & 0.883 \\ \addlinespace
    \multirow{4}{*}{GPT-3.5-turbo} & BERTScore    & 0.690 & 0.807  & \textbf{0.826} \\
                                & BLEURT    & 0.655 & 0.758  & \textbf{0.834}  \\
                                & ROUGE-L    & 0.706 & 0.767  & \textbf{0.817}  \\
                                & GPT-Judge& 0.683 & 0.891  & \textbf{0.912}  \\ \bottomrule
    \end{tabular}
\end{table}

For a more comprehensive analysis of the quality of the generated probes, we run an analysis to compare on the generated probes. As previously mentioned, for effective auditing, it is crucial that the generated probes not only remain relevant to the original question but also exhibit diversity within each question group. Consequently, we now proceed to evaluate both the relevance and diversity of these generated probes.

Initially, we assess the relevance of the generated probes by analyzing their semantic similarity and lexical overlap with the original questions. The findings are detailed in Table~\ref{tab:Result_RQ1_Similarity}. According to the table, the LLMAuditor model achieves the highest scores across all three LLMs, with GPT-3.5-turbo scoring the highest among the models for all similarity metrics. This performance suggests that the LLMAuditor method is particularly effective at generating probes that closely align with the original question, surpassing the other models. This high level of relevance can be attributed to the specially designed probe template used by LLMAuditor, which has been rigorously tested and refined through HIL evaluations. Incorporating HIL into the process will enhance the quality of the output, as highlighted in ~\citet{yang-etal-2023-human, zhang-etal-2023-human}.

\begin{table}[htbp]
    \centering
    \footnotesize
    \caption{Evaluating \textbf{Similarity} Between Original User Questions and Generated Probes.}
    \label{tab:Result_RQ1_Similarity}
    \begin{tabular}{@{}llccc@{}}
    \toprule
    LLM & Metric & \multicolumn{3}{c}{Baseline Models} \\ \cmidrule(l){3-5} 
        &       & Base Model & DSPy & LLMAuditor \\
        \hline \midrule
    \multirow{4}{*}{Falcon 7B} & BERTScore & 0.462  & 0.487  & \textbf{0.710} \\
                               & BLEURT    & 0.430 & 0.455 & \textbf{0.769} \\
                               & ROUGE-L    & 0.451 & 0.449 & \textbf{0.743} \\
                               \addlinespace
    \multirow{4}{*}{Llama2 7B} & BERTScore  & 0.498  & 0.502 & \textbf{0.712} \\
                                & BLEURT     & 0.460  & 0.493 & \textbf{0.730} \\
                                & ROUGE-L  & 0.499 & 0.515 & \textbf{0.768}  \\
                                \addlinespace
    \multirow{4}{*}{GPT-3.5-turbo} & BERTScore    & 0.510  & 0.526 &  \textbf{0.816} \\
                             & BLEURT       & 0.506  & 0.512 &  \textbf{0.791} \\
                            & ROUGE-L  & 0.518  & 0.517 &  \textbf{0.806} \\
                                     \bottomrule
    \end{tabular}
\end{table}

Similar to relevance, we also assess the diversity of the outputs by examining the dissimilarity of the probes within each question group. The results of this analysis are displayed in Table~\ref{tab:Result_RQ1_Disisimilarity}. It shows that the dissimilarity\footnote{Dissimilarity is calculated as: Dissimilarity = 1 - Similarity.} scores for LLMAuditor are lower in all three LLMs, and GPT-3.5-turbo achieve the lowest score, compared to those of other methods. This indicates that the LLMAuditor model is capable of generating more diverse probes for each question, in contrast to DSPy and the base model that utilizes the base probe template. This diversity is due to the especially human crafted probe template utilized by LLMAuditor. Here also we can see the effectiveness of having HIL and criteria is effective in generating output in a good quality, as shown in ~\citet{yang-etal-2023-human, zhang-etal-2023-human, 10.1145/3560815}.

\begin{table}[htbp]
    \centering
    \footnotesize
    \caption{Evaluating \textbf{Dissimilarity} Among Generated Probes for Each User Question.}
    \label{tab:Result_RQ1_Disisimilarity}
    \begin{tabular}{@{}llccc@{}}
    \toprule
    LLM & Metric & \multicolumn{3}{c}{Baseline Models} \\ \cmidrule(l){3-5} 
        &       & Base Model & DSPy & LLMAuditor \\
        \hline \midrule
    \multirow{4}{*}{Falcon 7B} & BERTScore & 0.861  & 0.793  & \textbf{0.236} \\
                               & BLEURT & 0.890  & 0.769 & \textbf{0.301} \\
                               & ROUGE-L  & 0.796 & 0.809  & \textbf{0.298} \\
                               \addlinespace
    \multirow{4}{*}{Llama2 7B} & BERTScore & 0.731 & 0.711  & \textbf{0.221} \\
                                & BLEURT  & 0.746 & 0.720 & \textbf{0.218}  \\
                                & ROUGE-L  & 0.707 & 0.743 & \textbf{0.230} \\
                                \addlinespace
    \multirow{4}{*}{GPT-3.5-turbo} & BERTScore & 0.729 & 0.733 & \textbf{0.153}     \\
                             & BLEURT & 0.718 & 0.706 & \textbf{0.202}  \\
                             & ROUGE-L   & 0.762   &  0.721 & \textbf{0.192}  \\
                                \bottomrule
    \end{tabular}
\end{table}

\subsection{Auditing LLMs (RQ2)}

The above discussion showed that LLMAuditor is superior in generating probs in compared to other manual and automatic probe generation methods. Now, we explore the overall performance of LLMAuditor in auditing LLMs for hallucinations and observe the results alongside those from another auditing tool, Adatest++~\cite{rastogi2023supporting}. 

Adatest++ is an LLM auditing tool that enables users to submit questions and evaluate the LLM's responses by manually labeling them as ``Fail'', ``Pass'', or ``Not sure''. In this evaluation, we use Adatest++ to audit hallucination in the TruthfulQA dataset. An annotator, trained according to the guidelines provided in the Adatest++ paper, assess the responses based on these categories (``Fail'', ``Pass'', or ``Not sure'').

We employ two evaluation metrics to compare the performance of Adatest++ and LLMAuditor. The first metric, ``Number of Fails", taken from~\cite{rastogi2023supporting}, tracks the instances where each model's response is marked as a ``Fail". For Adatest++, this count is directly based on the annotator's assessment. For LLMAuditor, we use the ``GPT\_Judge" metric from the TruthfulQA paper, which classify responses in as either hallucinated or not hallucinated with a threshold of 0.5\footnote{Following the threshold used in the original paper.}. LLMAuditor model presents five responses per question. If any of these responses is identified as hallucinated, we categorize the entire group as fail and direct the auditor to further investigate all the responses in that group. This structured approach allows for detailed comparison and evaluation of how each model handles potential hallucinations in responses.

With observing the results in Table~\ref{tab:Result_RQ2}, we note that the number of failures across both models and all three LLMs is consistent. For Adatest++, Falcon 7B recorded 340 answers marked as failures by the auditor. Similarly, the numbers for Llama2 7Band GPT-3.5-turbo are 371 and 314, respectively. These figures are based on the auditor's subjective judgment. In contrast, the number of failures for LLMAuditor, determined by the GPT\_Judge metric for classifying hallucinated answers, does not rely on subjective assessments. The respective failure counts for LLMAuditor are 331 for Falcon, 316 for Llama2, and 278 for GPT-3.5-turbo, which are comparable to those derived using the GPT\_Judge metric. Overall, the results of both the Adatest++ model and LLMAuditor align and shows that LLMAuditor is a reliable method for uncovering issues within LLMs.

\begin{table}[h]
    \centering
    \footnotesize
    \caption{Evaluation of auditing hallucination on TruthfulQA dataset across Falcon, Llama 2, and GPT-3.5-Turbo.}
    \label{tab:Result_RQ2}
    \begin{tabular}{@{}llcc@{}} 
    \hline
    \toprule
              & Model & \ Number of Fails & GPT-Judge  \\
    \hline
    \midrule
    \multirow{3}{*}{Adatest++} & Falcon 7B & 340 & 0.792 \\
                               & Llama2 7B & 371 & 0.831 \\
                               & GPT-3.5-turbo    & 314 & 0.859  \\ \addlinespace
    \multirow{3}{*}{LLMAuditor} & Falcon 7B& 331 & 0.865 \\
                               & Llama2 7B & 316 & 0.885 \\
                               & GPT-3.5-turbo    & 278 &  0.912 \\
    \hline
    \end{tabular}
    \label{tab:table4}
\end{table}

In summary, our analysis demonstrated that LLMAuditor is a reliable and automatic method for auditing inconsistencies LLMs and uncover disparities. Our findings highlight the importance of generating probes that are both relevant and diverse, which significantly aids in auditing inconsistencies within LLMs. Ideally, in a model devoid of any inconsistencies, we would expect LLMs to produce similar responses to similar prompts. However, if LLMs yield varying answers to comparable questions, it serves as an indication of potential inconsistencies that might be present in the models, signaling the need for more thorough auditing. LLMAuditor is capable of identifying these deviations, effectively flagging abnormal behavior in the outputs of LLMs.

\section{6 Conclusion} \label{6 Conclusion}

LLMs have evolved to become crucial tools for users seeking answers and assistance with tasks. As LLMs become more integrated into daily life, there is an increasing need for practical understanding of limitations such as bias \citep{Motoki2023, Talat2022, Thakur2023}, inconsistencies \citep{Tam2023, Ye2023}, and  hallucination \citep{chen2023hallucination, yang2023new}. These studies focus on addressing known issues, or {\em known unknowns}, through targeted audits of potential problems. However, they do not tackle {\em unknown unknowns}.

An effective method for evaluating LLMs regarding both known unknowns and unknown unknowns inconsistencies involves generating multiple variations of the same question. This approach elicits a range of responses, thereby testing the model's ability to consistently understand and interpret the underlying intent of the query. This technique involves rephrasing or altering the original question in several ways while maintaining its core intent. By presenting different formulations of the same query to the LLM, we can effectively audit the model's comprehension capabilities in response generation. This method is particularly useful for identifying various operational issues within the LLM. For instance, if the LLM responds differently to the same question posed in different ways, it may indicate certain underlying problems such as bias and hallucination. 

In this paper, we presented a systematic and generalizable framework, LLMAuditor, with a HIL component for auditing LLMs: (1) Probe Generation, which involves determining criteria to assess responses, and designing a prompt template to generate probes; and (2) Probe Answering, where LLMs generate answers for these probes.

The case study, which utilized questions from the TruthfulQA dataset~\cite{liu2023truthfulness}, demonstrates the feasibility of generating reliable and effective probes for auditing LLMs. This process involved creating multiple variations of the same question with HIL verification, each designed to audit different aspects of the LLM's capabilities. By employing this method, we could conduct a more thorough audit of the LLM’s performance, with each question variant probing different dimensions of the model's understanding and response generation.

The results from this case study underscore the importance of scientific rigor in the process of auditing LLMs. The use of varied question versions allowed us to uncover nuances in the LLM's responses that might not be evident when using a more uniform or singular approach. This methodology provides a comprehensive view of the LLM's performance across a spectrum of scenarios.

Note that while we applied our method of auditing to testing an LLM for potential issue of hallucination, the proposed framework is general enough for other forms of auditing. For instance, it is feasible for one to generate the auditing probes through LLM1 using different criteria than relevance and novelty. One such possibility is using personas to generate the probes, in which LLM1 is asked to create variations of the original question as it might be asked by a small set of different personas. This may be especially useful if LLM2 is being audited for potential biases based on who is asking the question or how they are posing the question. Similarly, while we chose to create five probes for our experiments, one could vary that to suite their needs. In essence, the auditing framework provided here is general enough for a number of different scenarios and needs for auditing.

While LLMAuditor provides a scalable solution for auditing LLM performance, it also comes with limitations. Although we have made progress in auditing LLMs for various aspects such as bias detection and fact-checking, we have not yet conducted a user study to evaluate the effectiveness of our method in terms of user satisfaction. Additionally, one of the restrictions is its narrow focus on question-and-answer tasks, which excludes other NLP tasks like translation and summarization. This limited scope may hinder the evaluation of LLM behavior beyond question-and-answer tasks. Moving forward, our aim is to expand the framework to encompass various NLP tasks in future studies.

\section{Ethical Considerations}

This study presents a framework for HIL validation of model-generated probes for LLM auditing. This approach ensures human involvement in assessing the data used to audit an LLM, potentially serving as a “sanity check” and ensuring that the content of the audit is aligned with human values. In this sense, it is intended to contribute to a more ethical approach to LLM development and deployment.

However, it should be noted that HIL approaches incur the risk of marginalization of data annotators ~\citep{10.1145/3411764.3445518}. We relied on seasoned researchers who are also co-authors of this paper. However, if external assessors are used, extra attention must be paid to have them treated fairly and compensated appropriately for their labor. This is especially important for crowdsourcing workers.

\vspace{1em}
\noindent
{\bf ACKNOWLEDGEMENTS}

\noindent
This work was supported by the Bill and Melinda Gates Foundation.

\clearpage 

\bibliography{Reference}

\end{document}